\documentclass{article}
\author{Rasoul Shahsavarifar, Jithu Chandran, Mario Inchiosa, Amit Deshpande, Mario Schlener, \\Vishal Gossain, Yara Elias, Vinaya Murali \vspace{.5cm}
\\\{rasoul.shahsavarifar, mario.schlener, vishal.gossain, yara.elias\}@ca.ey.com\\\{marinch, amitdesh\}@microsoft.com\\\{jithu.chandran, vinaya.murali\}@in.ey.com}
\title{\textbf{Identifying, measuring, and mitigating individual unfairness for supervised learning models and application to credit risk models\footnote{This paper is the second publication of the Microsoft-EY collaboration on the fairness of machine learning models. The first paper with the title of \href{https://www.microsoft.com/en-us/research/uploads/prod/2020/09/Fairlearn-EY_WhitePaper-2020-09-22.pdf}{"Assessing and mitigating unfairness in credit models with the Fairlearn toolkit"} which was focused on group fairness was published in 2020.\\Although this white paper touches on compliance with antidiscrimination and related laws, none of the guidance or recommendations expressed in this white paper should be taken as legal advice or as a substitute for working with legal professionals to ensure that AI systems comply with applicable laws.}}}
\date{\today}
\usepackage{hyperref}
\hypersetup{
    colorlinks=true,
    linkcolor=blue,
    citecolor= blue,      
    urlcolor=blue,
    pdfpagemode=FullScreen,
    }
\usepackage{fullpage}
\usepackage{amsmath}
\usepackage{enumerate}
\usepackage{url}
\usepackage{algorithm}
\usepackage{algorithmicx}
\usepackage[noend]{algpseudocode}
\usepackage{pifont}
\makeatletter
\def\BState{\State\hskip-\ALG@thistlm}
\makeatother
\usepackage{polynom}
\usepackage{multirow}
\usepackage{subfig}
\usepackage{amsfonts}
\usepackage[all]{xy}
\usepackage{graphicx}
\usepackage{titlesec}
\usepackage{color}
\usepackage{tabularx}
\usepackage{caption}
\usepackage{array}
\usepackage{algorithm}
\usepackage{pifont}
\makeatletter
\def\BState{\State\hskip-\ALG@thistlm}
\makeatother
\usepackage[english]{babel}
\usepackage{amsthm}
\usepackage[english]{babel}
\theoremstyle{definition}

\usepackage{amsopn}

\begin{document}
\maketitle
\vspace{1cm}
\section{Executive Summary}
\label{sec:ex_sum}
In the past few years, Artificial Intelligence (AI) has garnered attention from various industries including financial services (FS). AI has made a positive impact in financial services by enhancing productivity and improving risk management. While AI can offer efficient solutions, it has the potential to bring unintended consequences. One such consequence is the pronounced effect of AI-related unfairness and attendant fairness-related harms. These fairness-related harms could involve differential treatment of individuals; for example, unfairly denying a loan to certain individuals or groups of individuals.
\\\\The fairness of AI models can be studied considering two ways of measuring "fair" outcomes, group fairness and individual fairness. Individual fairness implies similar individuals are treated similarly, and group fairness implies that different groups based on sensitive features are treated similarly\footnote{In some recent publications, the group and individual fairness definitions are referred to as statistical measures of fairness and similarity-based measures of fairness, respectively.}.  For example, individual fairness implies that spouses with shared assets and similar finances would generally be offered similar credit limits. In contrast, group fairness implies that a metric such as the rate at which individuals are approved for loans is equal across genders. Achieving group fairness does not necessarily guarantee individual fairness and vice versa. There can be scenarios where a model achieves group fairness; however, it does not treat similar individuals similarly. The appropriate weighting of group and individual fairness should be based on the use case under consideration and the fairness priorities one aims to achieve. 
\\\\In this paper, we focus on identifying and mitigating individual unfairness and leveraging some of the recently published techniques in this domain, especially as applicable to the credit adjudication use case. We also investigate the extent to which techniques for achieving individual fairness are effective at achieving group fairness.
\\\\We have explored multiple approaches available in literature that can mitigate individual unfairness. In evaluating the approaches, special consideration must be given to the use of sensitive information (e.g., race, gender, age, etc.) because financial regulations around the world such as the US Equal Credit Opportunity Act (ECOA) prohibit the use of sensitive information in model training~\cite{CHAPTERSUBCHAPTER}. The prohibition pertains to using sensitive information in training the model to achieve improved model accuracy (e.g., to achieve better default predictions). The focus of this work is specifically on techniques wherein the use of sensitive information can be decoupled from the main model training as much as possible, and the sensitive information is not used for model prediction but only to collect and aggregate statistics during training. This ensures that the resultant model uses sensitive information in a clear and responsible manner, which increases adherence with the model risk guidelines of financial regulators.
\\\\Our main contribution in this work is functionalizing a two-step training process which involves learning a fair similarity metric from a group sense using a small portion of the raw data and training an individually "fair" classifier using the rest of the data where the sensitive features are excluded. The key characteristic of this two-step technique is related to its flexibility, i.e., the fair metric obtained in the first step can be used with any other individual fairness algorithms in the second step. Furthermore, we developed a second metric (distinct from the fair similarity metric) to determine how fairly a model is treating similar individuals. We use this metric to compare a "fair" model against its baseline model in terms of their individual fairness value. Finally, some experimental results corresponding to the individual unfairness mitigation techniques are presented, and the gender-wise accuracy, False Negative Rate (FNR), False Positive Rate (FPR), and Area Under Curve (AUC) of the models are also computed to see if the achieving individual fairness has any effect on group fairness.

\section{The Individual Fairness Problem and Its Importance} \label{sec:intro}
Algorithmic decisions by AI/ML systems can potentially induce and amplify multiple fairness-related harms to different sensitive demographics (e.g., race, gender) in terms of allocation of resources, quality of service, stereotyping, over- or under-representation in data as well as favourable outcomes etc~\cite{kelly2021algorithmic}. In financial applications such as loan adjudication, such harms could mean denial of access to funds in critical times for certain underrepresented minorities, unjustified stereotyping of some people as potential loan defaulters, and as a result, may carry a significant risk of legal and reputational damage for the lender. In many applications, the social and economic harms are due to a combination of multiple fairness-related harms listed above.
\\\\Algorithmic fairness can be classified into two categories, group fairness and individual fairness. Group fairness relates to fairness-related harms that affect groups of people, such as those defined in terms of sensitive features such as race, gender, age, or disability status. Group fairness is measured through standard group metrics such as accuracy, FPR, FNR, AUC, equality of opportunity or equality of error across various groups, etc. Definitions of group fairness metrics are widely accepted and are straightforward from a computational point a view. On the other hand, individual fairness draws from the notion that \textit{"similar individuals should be treated similarly"}. Unlike in group fairness, there are no standard definitions and metrics for individual fairness. The definition of what constitutes similarity between individuals is subjective and can have multiple interpretations. Hence, individual fairness has many variations based on the definition of the similarity metric chosen. 
Moreover, group and individual fairness principles may not be compatible, achieving one does not necessarily imply achieving the other.\textbf{We survey and compare current approaches in the literature and apply them to a credit risk adjudication use case}. 
\section{The Landscape of Individual Fairness Approaches }
\label{sec:landscape}
The first step in the process of finding an approach which can identify and mitigate individual unfairness is to understand the definition of similarity. We look for "similar" individuals who are treated differently, resulting in individual unfairness. At a high level, any fairness algorithm looks for unfairness instances and processes them in a way that the treatment given to the similar individuals by the new model ends up being similar. There are several ways with which researchers have approached the problem for individual unfairness and tried to mitigate it. This paper discusses a few of these techniques.
\subsection{Primary Works on Individual Fairness}
\label{sec:primary-work}
Before diving into the unfairness mitigation techniques surveyed in this paper, two methodologies which are considered as a basic framework are discussed below. Both these techniques define individual fairness based on the assumption that a fair metric exists.
\subsubsection{Perfect Metric-Fairness}
\label{sec:lipschitz}
\textit{Dwork et al.} proposed a similarity-based individual fairness notion in \cite{dwork2012fairness} in which a classifier is said to be individually fair if the distributions assigned to similar individuals are similar. For every arbitrary pair of individuals, the outcome of the classifier is individually fair if and only if the Lipschitz condition holds, that is, the classifier $h$ satisfies the $(D,d)-$Lipschitz property for every pair of individuals $x,x'$ in dataset $S$, we have
\[
D(h(x),h(x'))\leq(x,x'),
\]
where $d$ is the similarity metric, $d(x,x')\in [0,1]$ and $D$ is the similarity measure for distributions.
\subsubsection{Approximate Metric-Fairness}
\label{sec:approx_}
In~\cite{yona2018probably}, Rothblum and Yona presented a framework that yields a fair classifier that allows a small fraction of fairness error hence called Approximate Metric-Fairness. In this technique, a classifier h from a class H of probabilistic classifiers is Probably Approximately Correct and Fair (PACF), if it satisfies the following two conditions:
\begin{itemize}
\item[a.] Fairness: The predictor is approximately metric fair i.e., the probability of a pair of individuals being treated unfairly is bounded by a small fraction. In other words, metric-fairness holds for all but a small fraction $\alpha \in [0,1)$, of pairs of individuals.
\item[b.] 	Accuracy: For a certain subspace $H'$ of $H$, and for any fair classifier $h'\in H'$, the loss incurred by classifier $h$ is always smaller than that of $h'$.
\end{itemize}
Based on intuition of approximate metric fairness, some techniques such as Metric Free Individual Fairness Online Learning introduced in~\cite{bechavod2020metric} define fairness violations which are leveraged to achieve individual fairness. However, some other techniques such as Sensitive Subspace Robustness~\cite{yurochkin2019training} (SenSR) and Individually Fair Gradient Boosting \cite{vargo2021individually} (IFGB) base their notion of fair metric on the work of Dwork et. al. The next section discusses these and other techniques that are available to achieve some level of individual fairness.
\subsection{brief walkthrough of various techniques to progress towards individual fairness}
\label{sec:verios_techniques}
This section focuses on various techniques to progress towards individual fairness. We explore a few characteristics of these techniques, such as the following:
\begin{itemize}
\item whether a fair metric is required to be learned to establish similarity between individuals,
\item whether the technique achieves group fairness, too,
\item whether the model undergoes adversarial training, meaning the model is made robust to disturbances caused by unfair instances,
\item whether a human auditor is required during the process of training a fair model, and
\item whether the sensitive feature is one of the predictors in the fair model training.
\end{itemize}
Since the focus of this study is on the credit adjudication use case, of the characteristics listed above, the involvement of sensitive feature in training a fair model is of increased focus and scrutiny. That is because, in the credit adjudication domain, it is prohibited by regulatory bodies to consider the sensitive features as predictor features. Some approaches such as IGD~\cite{lohia2019bias}, Learning Fair Representation~\cite{zemel2013learning}, Pairwise Fair Representation~\cite{lahoti2019operationalizing}, and Cooperative Contextual Bandits \cite{hu2020metric}, cannot be used to work on a credit adjudication use case because these methods use the sensitive features in training their fair classifier.
\\\\On the other hand, the metric free methodologies (e.g.~\cite{bechavod2020metric}) are recently introduced in literature. In a metric-free approach, the sensitive features are excluded from the entire process of achieving individual fairness. The training phase of a metric-free method considered a human-in-the loop which makes this method not very practical specially for the real-world problems (e.g., credit adjudication) which usually deal with a large volume of data, and also the human biasness is potentially possible.
\\\\Some other methodologies such as SenSR~\cite{yurochkin2019training} and IFGB~\cite{vargo2021individually} suggest another way of achieving the individual fairness which does not need the involvement of human-in-the-loop and does not consider sensitive features in the training of the main model; hence, such methodologies do not contradict the regulatory requirements in the credit adjudication domain. In these techniques, the use of the sensitive information is separated from the main model training. This separation criterion is met using the two-step training process. An overview of the metric free and two-step training approaches is provided below.
\paragraph{Metric-Free Individual Fairness in Online Learning:}
This approach is an online learning problem with individual fairness as a constraint.  It leverages fairness violation instances to obtain a policy that minimizes misclassification loss and the number of fairness violations. This method requires an auditor-in-the-loop to identify a pair of individuals (per round) where fairness violation has occurred. Note that the auditor only comes up with one pair of violation. If there are multiple violations, the auditor chooses one arbitrarily. The goal is to achieve "sub-linear regret" which means that the loss due to misclassification and unfair instances are minimized with each passing round. This approach provides a general reduction framework that reduces the online fair classification to standard online classification with no fairness constraint. Reduction helps to leverage the existing standard online learning algorithms to simultaneously minimize cumulative classification loss and number of the fairness violations. To obtain this reduction framework synthetic data is created. This synthetic data contains certain number of copies of the violation pairs along with all the original data points. At every round t for a violation pair, this synthetic data is passed to an algorithm that returns a policy corresponding to $(t+1)^{th}$ round that has sub-linear regret.
\paragraph{Sensitive Subspace Robustness} 
The sensitive subspace robustness, also known as SenSR, is a two-step training approach which can be applied on smooth models, and it aims to build a model that is robust to any perturbations made by the sensitive features. The rationale behind SenSR can be explained as correspondence studies where the goal is to build a space comparable to the training data on which the model performs poorly.
\\\\Modifying the setup proposed by Dwork et. al. in~ \cite{dwork2012fairness}, this approach enforces the output of the model to be similar not only on all comparable individuals, but also to the training label. Thus, the accuracy of the model on the training data will be preserved while enforcing the individual fairness during training. Furthermore, this approach considers replacing the fair metric by the fair earth mover's distance \footnote{The earth mover distance, also known as Wasserstein distance or Kantorovich–Rubinstein metric is a distance function defined between probability distributions on a given metric space.}which enables the comparisons between the probability distributions instead of samples which helps to generalize this notion of individual fairness. 
\\\\In this approach, the performance of the model is assessed on the perturbed space by optimizing the loss incurred by the same. The fair model is trained with an adversarial approach where individual fairness violations are considered as adversarial attacks and the model is trained to be robust against such attacks. This methodology consists of two main steps: training a fair metric from a group sense and training an individually fair classifier. A small portion of the raw data is required to train the fair metric. For SenSR to satisfy the legal requirements of not using the sensitive feature in training the model, the small portion of the data set which is used to train the fair metric should be excluded from the training data of the classifier, and not be used in training the fair classifier. Further details can be found in Section \ref{sec:usage_sen_features}.
\paragraph{Individually Fair Gradient Boosting}
Another two-step training approach is the individually fair gradient boosting algorithm, also known as IFGB, which operationalizes individual fairness on smooth and non-smooth models such as the decision trees. This method is obtained from making some modifications to the concept of SenSR. Instead of building a comparable space same as the one in SenSR, IFGB looks for comparable samples from the input dataset where the model performs poorly. Hence to obtain a fair model, this approach solves a finite-dimensional linear program which is based on the loss values. The idea of restricting to the comparable sample from the input dataset is adopted since it is hard to solve the (distributional robustness) optimization for a non-smooth model.  However, the rationale behind learning a fair metric in IFGB is similar to that of SenSR.  The IFGB methodology also consists of two main steps: training a fair metric from a group sense and training an individually fair classifier. The requirements and conditions for training the fair metric and fair classifiers are the same as in the SenSR approach. 
\section{Our Approach}
\label{sec:approach}
As explained in Section \ref{sec:landscape}, there are multiple notions of individual fairness. The notion of fairness that one would consider for modelling could vary based on the use case. Broadly, information regarding fairness can be encoded into the training using two approaches, either it can be learned from the data, or it can be explicitly supplied during training in the form of labels indicating pairs of individuals that are similar or dissimilar. However, having enough pairwise labelled dataset for training is difficult to achieve for credit adjudication use cases. This is mainly because for data labelling, each pair of individuals must then be separately evaluated for similarity or dissimilarity. Moreover, curating such a dataset would require significant human judgement, which could inadvertently introduce further unfairness. Hence, our proposed approaches for the credit adjudication use case encode the notion of individual fairness by learning the fairness information from the data itself. 
\\\\The fair learning methodologies that we propose for credit adjudication use case are SenSR and IFGB (See Section \ref{sec:verios_techniques}). SenSR and IFGB consist of a two-step training process. First step entails learning a fair metric from a group sense and the second step consists of using the fair metric for adversarial training. Both the approaches are similar, except that SenSR can only be applied to smooth models (i.e., differentiable with respect to the inputs). IFGB can be applied to both smooth models (e.g., neural networks) and non-smooth models (e.g., XGBoost). 
In comparison to SenSR, IFGB is limited in terms of the span that the perturbed inputs\footnote{Perturbed inputs are similar to the original input (in a fair sense) but one that also flips the decision of the model. The perturbed inputs are required in the adversarial fairness training methods e.g. IFGB.} take. This is because, the perturbed inputs in IFGB can only be one of the other inputs from the training set. However, in SenSR the perturbed input can span the entire input space.
\subsection{Usage of Sensitive Information and Conformity with Compliance Laws}
\label{sec:usage_sen_features}
Access to sensitive information in banking is highly restricted. Additionally, regulations prohibit the use of sensitive information in model training to improve model accuracy. However, as we discussed in Section 3.2, excluding the sensitive information from the entire process of individual unfairness mitigation requires employing a metric-free approach. As mentioned in Section 3.2, the metric-free approaches are not very practical for the credit adjudication problem. Hence, to achieve individual fairness that satisfies regulatory requirements, one practical approach is to separate the use of the sensitive information from the main model training. This separation criterion is met for SenSR and IFGB using a two-step training process. In the first step, a fair metric is learned from a group sense by using a small portion of the raw data where the sensitive information is included. Such portion of the data will be removed from the dataset thereafter.  In the second step, an individually fair model is trained using the fair metric and the remaining portion of the dataset where the sensitive information is completely excluded. The usage of the learned fair metric does not require sensitive information. The fair distance metric provides a distance measure between two data points (without the sensitive information) such that the value is smaller if the corresponding two individuals are similar and vice versa. The parameters of the fair metric are held constant during the final model training.
\\\\Therefore, SenSR and IFGB provide better chance to achieve regulatory compliance compared to other available methods because they do not directly use the sensitive features in the final model training. Moreover, addressing regulatory concern that the use of sensitive information would introduce unfairness in the model, SenSR and IFGB use the sensitive information in an adversarial setting thereby mitigating unfairness.
\section{Empirical Analysis}
\label{sec:emp_analysis}
We implement SenSR and IFGB algorithms on a real-world credit default dataset. The objective is to use these algorithms to mitigate individual unfairness present in the baseline model. Baseline models considered in the case of SenSR and IFGB are neural networks and XGBoost respectively. Performance of the unfairness mitigation algorithm is evaluated based on the trade-off between the fairness improvement and performance reduction (upon fairness treatment) as compared to the baseline model. As another set of experiments, we calculated the gender-wise accuracy, FPR, FNR, and AUC for the SenSR and IFGB, and for their corresponding baseline models to confirm whether the group fairness is also achieved.
\subsection{Data}
\label{sec:data}
The credit default dataset considered in the analysis consists of a total of $52,588$ records and $29$ features. This dataset is a subset of a publicly available dataset on Kaggle\footnote{\href{https://www.kaggle.com/c/home-credit-default-risk}{Home credit default risk dataset.}}. The training set consist of $31,646$ records and the testing set consists of $20,942$ records. The training set is further divided for the purpose of main model training and the fair distance metric training. The further splitting of the training set is done to ensure that the dataset used for fair metric learning, which requires sensitive information (i.e., Gender), is separated from the main model training (see Section \ref{sec:usage_sen_features}). Table \ref{tbl:data} provides the details of the data splits.

\begin{table}[!ht]
\caption{Data Description}
\label{tbl:data}
\vspace{-.5cm}
\begin{center}
\begin{tabular}{|c |c |c |c |}
\hline
{} & \multicolumn{2}{c|}{\textbf{Training}} &\multirow{2}{*}{\textbf{Testing}}\\
\cline{2-3}
 & {\textbf{Main Model}}&{\textbf{Fair Metric}} & \\
\hline
 \#\textbf{Records} & $23,145$ & $8,501$ & $20,942$ \\
\hline
 \#\textbf{Defaults}&$11,499$ &$4,324$ &$3,942$ \\
\hline
\end{tabular}
\end{center}
\end{table}

\subsection{Experimental Results}
\label{sec:exp-result}
Section \ref{sec:performance} provides the results of different performance metrics for the SenSR, IFGB, and their corresponding baseline models. Section \ref{sec:mitigation-metric} provides the results of the unfairness mitigation achieved in the models.
\subsubsection{Model Performance}
\label{sec:performance}
\begin{figure}[!ht]
  \centering
    \includegraphics[width=0.9\textwidth]{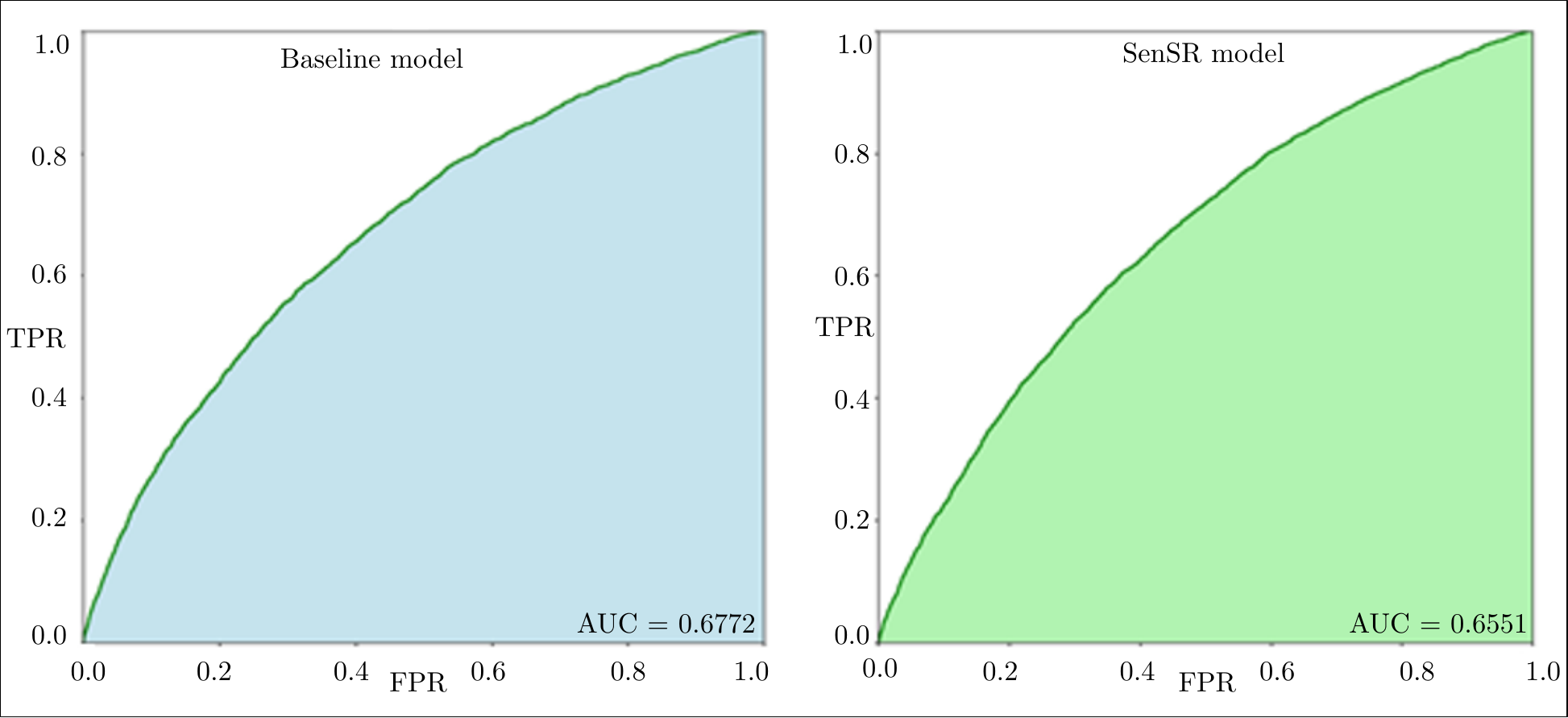}
  \caption{ROCs and AUCs of SenSR and its corresponding baseline model}
  \label{fig:sensr-roc}
\end{figure}

Table \ref{tbl:performence-sensr} and Figure \ref{fig:sensr-roc} provide the performance metric comparison between SenSR and its corresponding baseline neural network model.

\begin{table}[!ht]
\caption{Performance metrics of SenSR vs its baseline model}
\label{tbl:performence-sensr}
\vspace{-.5cm}
\begin{center}
\begin{tabular}{|c|c|c|c|c|}
\hline
\textbf{Data} & \textbf{Accuracy}& \textbf{FPR}&\textbf{FNR}&\textbf{AUC}    \\
\hline
\textbf{Baseline}& 63.92\%  & 35.36\% & 39.14\% & 0.6772 \\
\hline
\textbf{SenSR}&62.30\%  & 37.62\% & 38.03\% & 0.6551 \\
\hline
\end{tabular}
\end{center}
\end{table}

\begin{figure}[!ht]
  \centering
    \includegraphics[width=0.9\textwidth]{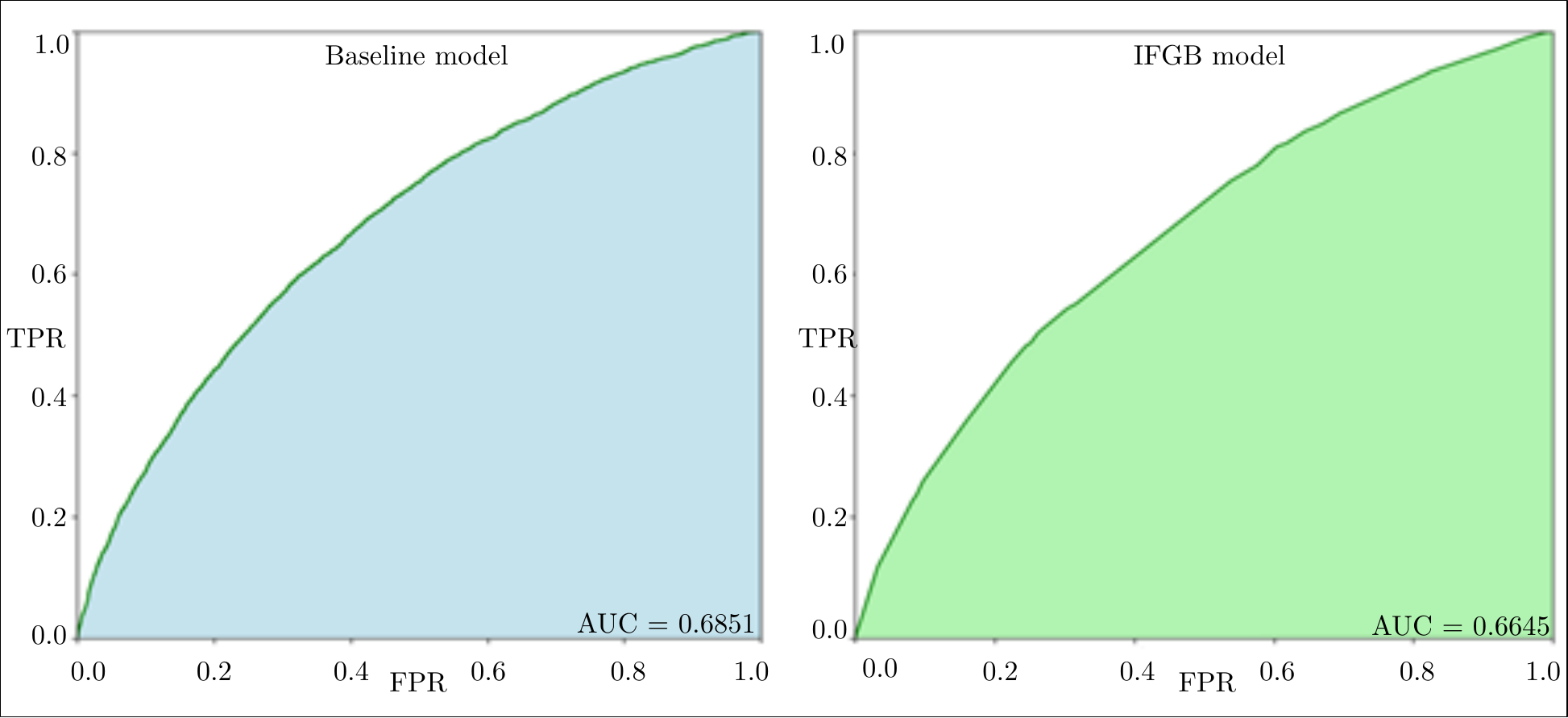}
  \caption{ROCs and AUCs of IFGB and its corresponding baseline model}
  \label{fig:ifgb-roc}
\end{figure}

Table \ref{tbl:performence-ifgb} and Figure \ref{fig:ifgb-roc} provide the performance metric comparison between IFGB and its corresponding baseline XGBoost model. Regarding the accuracy values for IFGB and baseline models in this table, it should be mentioned that it is not preordained that accuracy should decrease with increased fairness \cite{dutta2020there}. The observed increase in the accuracy but decrease in the AUC for IFGB compared to its baseline model can be related to using a highly unbalanced labelled dataset. The dataset consists of two classes $0$ and $1$ with $91.9\%$ and $8.1\%$ of the data, respectively. 

\begin{table}[!ht]
\caption{Performance metrics of IFGB vs its baseline model}
\label{tbl:performence-ifgb}
\vspace{-.5cm}
\begin{center}
\begin{tabular}{|c|c|c|c|c|}
\hline
\textbf{Data} & \textbf{Accuracy}& \textbf{FPR}&\textbf{FNR}&\textbf{AUC}    \\
\hline
\textbf{Baseline}& 64.74\%  & 34.48\% & 38.65\% & 0.6851 \\
\hline
\textbf{IFGB}&66.68\%  & 30.45\% & 45.66\% & 0.6645 \\
\hline
\end{tabular}
\end{center}
\end{table}

From the above results it can be observed that application of unfairness mitigation algorithm results in a reduction in the most of model performance results. 
\\\\Tables~\ref{tbl:performence-sensr-m} and~\ref{tbl:performence-sensr-f} present the values of gender-wise metrics which measure the performance of SenSR and its corresponding baseline model. Table~\ref{tbl:performence-sensr-diff} shows the differences between the related values in Tables~\ref{tbl:performence-sensr-m} and~\ref{tbl:performence-sensr-f}. 

\begin{table}[!ht]
\caption{Performance metrics of SenSR vs its baseline model for group Male}
\label{tbl:performence-sensr-m}
\vspace{-.5cm}
\begin{center}
\begin{tabular}{|c|c|c|c|c|}
\hline
\textbf{Data} & \textbf{Accuracy}& \textbf{FPR}&\textbf{FNR}&\textbf{AUC}    \\
\hline
\textbf{Baseline}& 61.68\%  & 38.88\% & 36.43\% & 0.6705 \\
\hline
\textbf{SenSR}&60.05\%  & 41.07\% & 36.26\% & 0.6375 \\
\hline
\end{tabular}
\end{center}
\end{table}

\begin{table}[!ht]
\caption{Performance metrics of SenSR vs its baseline model for group Female}
\label{tbl:performence-sensr-f}
\vspace{-.5cm}
\begin{center}
\begin{tabular}{|c|c|c|c|c|}
\hline
\textbf{Data} & \textbf{Accuracy}& \textbf{FPR}&\textbf{FNR}&\textbf{AUC}    \\
\hline
\textbf{Baseline}& 65.15\%  & 33.60\% & 41.22\% & 0.6770 \\
\hline
\textbf{SenSR}&63.53\%  & 35.90\% & 39.38\% & 0.6587 \\
\hline
\end{tabular}
\end{center}
\end{table}

\begin{table}[!ht]
\caption{Differences between metrics across gender for SenSR and its baseline model (Male-Female)}
\label{tbl:performence-sensr-diff}
\vspace{-.5cm}
\begin{center}
\begin{tabular}{|c|c|c|c|c|}
\hline
\textbf{Data} & \textbf{Accuracy diff}& \textbf{FPR diff}&\textbf{FNR diff}&\textbf{AUC diff}\\
\hline
\textbf{Baseline}& -3.47\%  & 5.28\% & -4.79\% & -0.0065 \\
\hline
\textbf{SenSR}&-3.48\%  & 5.17\% & -3.12\% & -0.0212 \\
\hline
\end{tabular}
\end{center}
\end{table}

Tables~\ref{tbl:performence-ifgb-m} and~\ref{tbl:performence-ifgb-f} provide the values of gender-wise metrics which measure the performance of IFGB and its corresponding baseline model. Table~\ref{tbl:performence-ifgb-diff} shows the differences between the related values in Tables~\ref{tbl:performence-ifgb-m} and~\ref{tbl:performence-ifgb-f}. As can be seen in Table~\ref{tbl:performence-sensr-diff}, the predictive equality (i.e., the differences in FPR across different groups) has not changed remarkably. For the baseline model, the predictive equality is $5.28\%$, and for the SenSR model, it is $5.17\%$ which shows a difference of $0.11\%$. After further test and considering other values in Table~\ref{tbl:performence-sensr-diff}, it seems that mitigating the individual unfairness using the SenSR approach does not worsen the group fairness.

\begin{table}[!ht]
\caption{Performance metrics of IFGB vs its baseline model for group Male}
\label{tbl:performence-ifgb-m}
\vspace{-.5cm}
\begin{center}
\begin{tabular}{|c|c|c|c|c|}
\hline
\textbf{Data} & \textbf{Accuracy}& \textbf{FPR}&\textbf{FNR}&\textbf{AUC}    \\
\hline
\textbf{Baseline}& 62.35\%  & 38.20\% & 35.85\% & 0.6787 \\
\hline
\textbf{IFGB}&63.65\%  & 34.41\% & 42.81\% & 0.6577 \\
\hline
\end{tabular}
\end{center}
\end{table}

\begin{table}[!ht]
\caption{Performance metrics of IFGB vs its baseline model for group Female}
\label{tbl:performence-ifgb-f}
\vspace{-.5cm}
\begin{center}
\begin{tabular}{|c|c|c|c|c|}
\hline
\textbf{Data} & \textbf{Accuracy}& \textbf{FPR}&\textbf{FNR}&\textbf{AUC}    \\
\hline
\textbf{Baseline}& 66.05\%  & 32.63\% & 40.64\% & 0.6847 \\
\hline
\textbf{IFGB}&68.34\%  & 28.47\% & 47.85\% & 0.6634 \\
\hline
\end{tabular}
\end{center}
\end{table}

\begin{table}[!ht]
\caption{Differences between metrics across gender for IFGB and its baseline model (Male-Female)}
\label{tbl:performence-ifgb-diff}
\vspace{-.5cm}
\begin{center}
\begin{tabular}{|c|c|c|c|c|}
\hline
\textbf{Data} & \textbf{Accuracy diff}& \textbf{FPR diff}&\textbf{FNR diff}&\textbf{AUC diff}\\
\hline
\textbf{Baseline}& -3.70\%  & 5.57\% & -4.79\% & -0.0060 \\
\hline
\textbf{IFGB}&-4.69\%  & 5.94\% & -5.04\% & -0.0057 \\
\hline
\end{tabular}
\end{center}
\end{table}

Considering the results in Table~\ref{tbl:performence-ifgb-diff}, and also performing further tests, it seems that mitigating the individual unfairness using the IFGB approach does not worsen the group fairness.
\\\\Motivated by these experimental results, one can consider a two-phase unfairness mitigation approach to achieve a model which treats all groups and Individuals fairly. The first phase of this approach can be considering appropriate group unfairness identification and mitigation techniques, and for the second phase, the SenSR or IFGB methods can be applied on the outcome of phase one. The reason that we suggest mitigating group unfairness in the first phase is that applying group fairness metrics and methodologies are more straightforward. Furthermore, both SenSR and IFGB are two-step methods that require to learn a fair metric from a group sense in their first step.

\subsubsection{Unfairness Mitigation Metric}
\label{sec:mitigation-metric}  
\begin{figure}[!ht]
  \centering
    \includegraphics[width=0.9\textwidth]{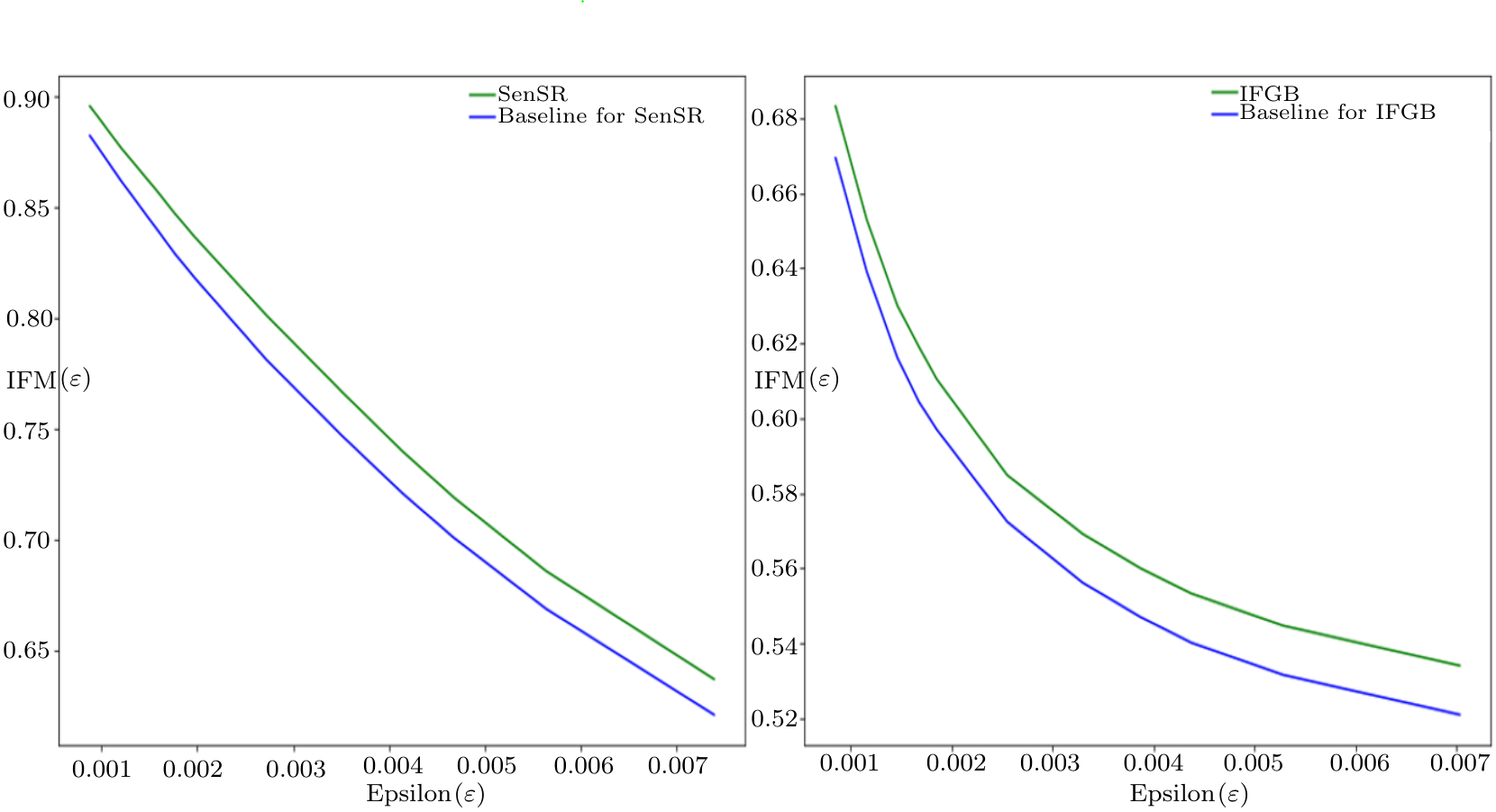}
  \caption{Fairness metric for SenSR and IFGB vs their corresponding baseline models}
  \label{fig:fairness-metric}
\end{figure}

Equation~\eqref{eq:ifm} provides the Individual Fairness Metric (IFM) which is used to quantify the individual fairness of a model. The metric computes the percentage of similar individuals that are treated similarly (i.e., that have the same model outcome). Similarity is measured based on the fair distance metric learned in the training phase.

\begin{equation}
\label{eq:ifm}
\frac{\#\text{similar individuals with same predicted responses by the model}}{\#\text{similar individuals}}
\end{equation}
Customising the typical definition of Recall, Equation~\eqref{eq:ifm} can be given by

\begin{equation}
\label{eq:ifm2}
IFM(\varepsilon)=\frac{\sum_{(x,x')\in S}I\left((d(x,x')\leq \varepsilon)\cap (h(x) = h(x'))\right)}{\sum_{(x,x')\in S}I(d(x,x')\leq \varepsilon)},
\end{equation}
where $I$ is an indicator function, $\varepsilon$ is the similarity threshold, and the other components of \eqref{eq:ifm2} are introduced in Section~\ref{sec:lipschitz} (\textit{Lipschitz property}).
\\\\Figure~\ref{fig:fairness-metric} provides the individual fairness metric (computed as shown in Equation~\eqref{eq:ifm2}) for SenSR, IFGB and their corresponding baseline models.
\\\\As can be seen in Figure~\ref{fig:fairness-metric}, the individual fairness metric computed for SenSR and IFGB are higher than that of their corresponding baseline models. Moreover, it can be observed that the metric decreases as $\varepsilon$ increases, which is expected. As presented in Tables~\ref{tbl:performence-sensr-m},~\ref{tbl:performence-sensr-f},~\ref{tbl:performence-ifgb-m}, and~\ref{tbl:performence-ifgb-f}, we have observed that the baseline Neural Network and XGboost models have comparable accuracy, FPR, FNR, and AUC across different groups; however, as illustrated in Figure~\ref{fig:fairness-metric}, they show significant differences in terms of individual fairness metric values.
\section{Model Monitoring}
\label{sec:monitoring}
Model monitoring procedure can consist of multiple factors such as developing the Key performance Indicator (KPI) guidelines, maintaining a well-prepared documentation (which includes model overrides, back testing, sensitivity testing, and data drift considerations) over time, group and individual fairness criteria, and model explainability considerations. Regarding the model fairness criteria, the enterprises should be able to enforce fairness in their model's outcome by analyzing transactions in production and finding biased behaviour by the model. There should be appropriate techniques in place in order to pinpoint the source of unfairness and actively mitigate the unfairness found in production environment. Regarding the fairness consideration, an efficient model monitoring procedure should
\begin{itemize}
\item automatically recommend common sensitive features to be monitored during the production.
\item consider detecting unfairness at runtime to catch impacts on business applications and compliance requirements without performing a time-consuming manual data analysis.
\item consider mitigating biases at runtime to enforce regulatory or enterprise fairness guardrails in real time. 
\end{itemize}

\section{Conclusion}
\label{sec:conclu}
In this paper, we presented several approaches to identify and mitigate individual unfairness and applied them to a credit adjudication model. Since the regulations of credit adjudication prohibit the modellers from using the sensitive features (e.g., gender, race, age) as a predictor variable, we have focused on the algorithms wherein the sensitive information is not used in the training of the models. In fact, the use of sensitive information is decoupled from the main model training as much as possible, and the sensitive information is not used for model prediction but only to collect and aggregate group-wise statistics during training. This objective is achieved through leveraging SenSR and IFGB methods to functionalize a two-step training process which involves training a fair metric from a group sense and training an individually fair model using two separate portions of the dataset. The key characteristic of this two-step training technique is related to its flexibility i.e., the fair metric obtained in the first step can be used with any other fairness algorithms in the second step. Finally, we performed some experiments to ensure the effectiveness of this technique. The experimental results suggest that the individual fairness of the credit adjudication model can be achieved without worsening the group fairness.
\\\\As a direction for the future work, the two-step training technique that we functionalized in this paper can be leveraged in other use cases in financial industry, healthcare, etc. Furthermore, a framework can be designed for both smooth and non-smooth models that let all individual fairness methodologies such as SenSR, IFGB, etc. to be applied on the same baseline; hence, the performance can be compared to one another. Generalizing such framework to include the individual fairness of unsupervised models is another direction for our future work in the area of model fairness.
\section{Acknowledgement}
\label{sec:acknowledgement}
The Authors would like to acknowledge \textit{Hanna Wallach} and \textit{Amit Sharma} from Microsoft Research and \textit{Hiwot Tesfaye} from Microsoft’s Office for Responsible AI for reading this paper carefully and providing us with detailed and very valuable comments.
\pagebreak

\bibliographystyle{plain}
\bibliography{individual_fairness}

\end{document}